\documentclass[runningheads]{llncs}
\usepackage{graphicx}
\usepackage{tikz}
\usetikzlibrary{shapes}
\usepackage{pgf}
\usepackage{pgfplots}
\pgfplotsset{compat=1.14}
\usetikzlibrary{arrows}
\usepackage{tkz-graph}
\usetikzlibrary{shapes.multipart}
\usetikzlibrary{calc}

\usepackage{hyperref}

\usepackage{array}
\newcolumntype{P}[1]{>{\centering\arraybackslash}p{#1}}
\usepackage{multirow}

\usepackage{float}

\newcommand{\ee}{end-to-end }

\begin{document}
%

\title{TED-LIUM 3: Twice as Much Data and Corpus Repartition for Experiments on Speaker Adaptation}

\titlerunning{TED-LIUM 3 corpus}
%
\author{Fran\c{c}ois Hernandez\inst{1} \and
Vincent Nguyen\inst{1} \and Sahar Ghannay\inst{2} \and 
Natalia~Tomashenko\inst{2} \and Yannick Est\`eve\inst{2}}
\authorrunning{F. Hernandez et al.}
%
\institute{Ubiqus, Paris, France \email{  flast@ubiqus.com}\\
\url{https://www.ubiqus.com} \and
LIUM, University of Le Mans, France \email{  first.last@univ-lemans.fr}\\
\url{https://lium.univ-lemans.fr/}}
\maketitle              
\begin{abstract}
In this paper, we present TED-LIUM release 3 corpus\footnote{TED-LIUM 3 is available on \url{https://lium.univ-lemans.fr/ted-lium3/}} dedicated to speech recognition in English, which multiplies the available data to train acoustic models in comparison with \mbox{TED-LIUM~2}, by a factor of more than two.
We present the recent development on Automatic Speech Recognition~(ASR) systems in comparison with the two previous releases of the TED-LIUM Corpus from 2012 and 2014.
We demonstrate that, passing from 207 to 452 hours of transcribed speech training data is really more useful for end-to-end ASR systems than for HMM-based state-of-the-art ones. This is the case even if the HMM-based ASR system still outperforms the end-to-end ASR system when the size of audio training data is 452 hours, with a Word Error Rate (WER) of 6.7\% and 13.7\%, respectively.
Finally, we propose two repartitions of the TED-LIUM release 3 corpus: the \textsl{legacy} repartition that is the same as that existing in release 2, and a new repartition, calibrated and designed to make experiments on \textsl{speaker adaptation}. Similar to the two first releases, TED-LIUM~3 corpus will be freely available for the research community.
\end{abstract}
\section{Introduction}
In May 2012 and May 2014, the LIUM team released two versions (respectively 118 hours of audio and 207 hours of audio) from the TED conference videos which were since widely used by the ASR community for research purposes. These corpora were called TED-LIUM, release 1 and release 2, presented respectively in~\cite{rousseau2012ted} and \cite{rousseau2014enhancing}. 
Ubiqus joined these efforts to pursue the improvements both from an increased data standpoint, as well as from a technical achievement one. We believe that this corpus has become a reference and will continue to be used by the community to improve further the results.
In this paper, we present some enhancements regarding the dataset, by using a new engine to realign the original data, leading to an increased amount of audio/text, and by adding new TED talks, which combined with the new alignment process, gives us 452 hours of aligned audio. A new data distribution is also proposed that is more suitable for experimenting with speaker adaptation techniques, in addition to the \textsl{legacy} distribution already used on TED-LIUM release 1 and 2.
Section~\ref{sec:T3} gives details about the new TED-LIUM 3 corpus.
We  present experimental results with different ASR architectures, by using Time Delay Neural Network (TDNN)~\cite{Peddinti2015ATD} and Factored TDNN (TDNN-F) acoustic models~\cite{Povey2018TDNNF} on the \textsl{legacy} distribution of TED-LIUM 3 in section~\ref{sec:HMMdesc}, and also exploring the use of a pure neural end-to-end system in section~\ref{sec:e2eexp}.
In section~\ref{sec:SAexp}, we report experimental results obtained on the \textsl{speaker adaptation} distribution by exploiting GMM-HMM and TDNN-Long Short-Term Memory (TDNN-LSTM)~\cite{peddinti2018low} acoustic models and two standard adaptation techniques (i-vectors and feature space maximum linear regression (fMLLR)). 
%
The final section is dedicated to discussion and conclusion.

\section{TED-LIUM 3 Corpus Description}
\label{sec:T3}
\subsection{Data, Alignment and Filtering}
\label{subsec:align}
All raw data (acoustic signals and closed captions) were extracted from the TED website. For each talk, we built a \texttt{sphere} audio file, and its corresponding transcript in \texttt{stm} format. The text from each \texttt{.stm} file was automatically aligned to the corresponding \texttt{.sph} file using the Kaldi toolkit \cite{Povey_ASRU2011}. This consists of the adaptation of existing scripts \footnote{\url{https://github.com/kaldi-asr/kaldi/blob/master/egs/wsj/s5/steps/cleanup/segment_long_utterances.sh}}, intended to first decode the audio files with a biased language model, and then align the obtained \texttt{.ctm} file with the reference transcript. To maximize the quality of alignments, we used our best model (at the time of corpus preparation) trained on the previous release of the TED-LIUM corpus. This model achieved a WER of 9.2\% on both development and test sets without any rescoring.
This means the ratio of aligned speech versus audio from the original 1,495 talks of releases 1 and 2 has changed, as well as the quantity of words retained. It increased the amount of usable data from the same basis files  by around 40\% (Table \ref{T2align}). In the previous release, aligned speech represented only around 58.9\% of the total audio duration (351 hours). With these new alignments, we now cover around 83.0\% of audio.

\begin{table}[h]
\centering
\caption{Maximizing alignments - TED-LIUM release 2 talks.}\label{T2align}
\begin{tabular}{|@{\hspace{1em}} m{3cm} @{\hspace{1em}}| P{2cm} | P{2cm} | P{2cm} |}
\hline
\multirow{2}{4em}{Characteristic} & \multicolumn{2}{c|}{Alignments} & \multirow{2}{4em}{Evolution} \\
\cline{2-3}
 & Original & New & \\
\hline
Speech & 207h & 290h & 40.1\% \\
\hline
Words & 2.2M & 3.2M & 43.1\% \\
\hline
\end{tabular}
\end{table}

A first set of experiments was conducted to compare equivalent systems trained on the two sets of data (Table \ref{T2alignRes}). With strictly equivalent models, there is no clear improvement of results for the proposed new alignments. Yet, there is no degradation of performance either. We will show in further experiments that the increased amount of data will not just be harmless, but also useful.

\begin{table}[h]
\centering
\caption{Comparison of training on original and new alignments for TED-LIUM release 2 data (Experiments conducted with the Kaldi toolkit - details in Section \ref{sec:HMMdesc}).}\label{T2alignRes}
\begin{tabular}{| P{4cm} | P{1.5cm} | P{1.5cm} | P{1.5cm} | P{1.5cm} |}
\hline
\multirow{2}{8em}{Model (rescoring)} & \multicolumn{2}{c|}{Original - 207h} & \multicolumn{2}{c|}{New - 290h} \\
\cline{2-5}
 & Dev & Test & Dev & Test \\
\hline
HMM-GMM (none) & 19.0\% & 17.6\% & 18.7\% & 17.2\% \\
HMM-GMM (Ngram) & 17.8\% & 16.5\% & 17.7\% & 16.1\% \\
HMM-TDNN-F (none) & 8.5\% & 8.3\% & 8.2\% & 8.3\% \\
HMM-TDNN-F (Ngram) & 7.8\% & 7.8\% & 7.7\% & 7.9\% \\
HMM-TDNN-F (RNN) & 6.8\% & 6.8\% & 6.6\% & 6.7\% \\
\hline
\end{tabular}
\end{table}

\subsection{Corpus Distribution: Legacy Version}
\label{subsec:T3legacy}
The whole corpus is released as what we call a \textsl{legacy} version, for which we keep the same development and test sets as the first releases.
Table \ref{T3stats} summarizes the characteristics of text and audio data of the new release of the TED-LIUM corpus. Statistics from the previous and new releases are presented, as well as the evolution between the two. Additionally, we mention that aligned speech (including some noises and silences) represents around 82.6\% of audio duration (540 hours).

\begin{table}[h]
\centering
\caption{TED-LIUM 3 corpus characteristics.}\label{T3stats}
\begin{tabular}{|@{\hspace{1em}} m{3cm} @{\hspace{1em}}| P{2cm} | P{2cm} | P{2cm} |}
\hline
\multirow{2}{4em}{Characteristic} & \multicolumn{2}{c|}{Corpus} & \multirow{2}{4em}{Evolution} \\
\cline{2-3}
 & v2 & v3 & \\
\hline
Total duration & 207h & 452h & 118.4\% \\
\hspace{5mm}- Male & 141h & 316h & 124.1\% \\
\hspace{5mm}- Female & 66h & 134h & 103.0\% \\
\hline
Mean duration & 10m 12s & 11m 30s & 12.7\% \\
\hline
Number of unique speakers & 1242 & 2028 & 63.3\% \\
\hline
\multicolumn{4}{c}{ }\\[-2ex]
\hline
Number of talks & 1495 & 2351 & 57.3\% \\
\hline
Number of segments & 92976 & 268231 & 188.5\% \\
\hline
Number of words & 2.2M & 4.9M & 122.7\% \\
\hline
\end{tabular}
\end{table}

\subsection{Corpus Distribution: Speaker Adaptation Version}\label{sec:adapt-corp}
Speaker adaptation of acoustic models (AMs) is an important mechanism  to reduce the
mismatch between the AMs and test data from a particular speaker, and today
it is still a very active research area. In order to design a suitable corpus for exploring  speaker adaptation algorithms, additional   factors and dataset characteristics, such as number of speakers, amount of pure speech data per speaker, and others,  should be taken into account.
In this paper, we also propose and describe the training, development and test datasets specially designed for the speaker adaptation task. These datasets are obtained from the proposed TED-LIUM 3 training corpus, but the development and test sets are more balanced and  representative  in characteristics (number of speakers, gender,
duration) than the original sets and more suitable for speaker adaptation
experiments. In addition, for the development and test datasets we chose only  speakers who are not present in the training data set in other talks.
The statistics for the proposed data sets are given in Table~\ref{tab_adapt_sets}.
\begin{table}[h]
\centering
\caption{\label{tab_adapt_sets}{Data sets statistics for the speaker adaptation task. Unlike the other tables, the duration is calculated only for pure speech (excluding silence, noise, etc.).}}
\centerline{\renewcommand{\tabcolsep}{3.5mm}
		\begin{tabular}{|c|c|c|c|c|}
			\hline
			 \multicolumn{2}{|c|}{\multirow{2}{*}{Characteristic}}  & \multicolumn{3}{c|}{Data set} \\ \cline{3-5}
			 \multicolumn{2}{|c|}{} &  Train & Dev. & $\mathrm{Test}$  \\
			\hline
			\multirow{3}{*}{\shortstack[c]{Duration of speech, \\ hours}} &  Total & 346.17    & 3.73 & 3.76 \\
			&Male           & 242.22  & 2.34  & 2.34  \\
			&Female          & 104.0    &  1.39  & 1.41   \\ \hline 
			\multirow{3}{*}{\shortstack[c]{Duration of speech \\ per speaker, minutes}} & Mean   &  10.7 & 14.0 &  14.1    \\ 
			& Min. &  1.0  &  13.6  & 13.6    \\ 
			& Max.  & 25.6 &  14.4  & 14.5   \\ \hline 
			\multirow{3}{*}{Number  of speakers} & Total  & 1938   & 16  & 16   \\
			& Male                & 1303     & 10  & 10   \\
			& Female              & 635    &  6  & 6   \\ \hline 
		Number of words & Total & 4437K
 & 47753 & 43931  \\ \hline 
 Number of talks & Total & 2281
 & 16 & 16  \\ \hline
\end{tabular}
}
\end{table}

\section{Experiments with State-of-the-art HMM-based ASR System}
\label{sec:HMMdesc}
We conducted a first set of experiments on the TED-LIUM release 2 and 3 corpora using the Kaldi toolkit. These experiments were based on the existing recipe \footnote{\url{https://github.com/kaldi-asr/kaldi/tree/master/egs/tedlium/s5_r2}}, mainly changing model configurations and rescoring strategies. We also kept the lexicon from the original release, containing 159,848 entries. For this, and all other experiments in this paper, no \textsl{glm} files were applied to deal with equivalences between word spelling (\textsl{e.g.} doctor \textsl{vs.} dr).
\subsection{Acoustic Models}
All experiments were conducted using chain models \cite{Povey2016PurelySN} with the now well-known TDNN architecture~\cite{Peddinti2015ATD} as well as the recent TDNN-F architecture~\cite{Povey2018TDNNF}. Training audio samples were randomly perturbed in speed and volume during the training process. This approach is commonly called \textsl{audio augmentation} and is known to be beneficial for speech recognition~\cite{ko2015audio}.
\subsection{Language Model}
Two approaches were used, both aiming at rescoring lattices. The first one is an N-gram model of order 4 trained with the \textsl{pocolm} toolkit\footnote{\url{https://github.com/danpovey/pocolm}}, which was pruned to 10 million N-grams. We also considered a RNNLM with letter-based features and importance sampling \cite{Xu2017NeuralNL}, coupled with a pruned approach to lattice-rescoring \cite{Xu2017APR}. The RNNLM we retained was a mixture of three TDNN layers with two interspersed LSTMP layers \cite{Sak2014LongSM} containing around 10 million parameters. The latter helps to reduce the word error rate drastically.
We used the same corpus and vocabulary in both methods, which are those released along with TED-LIUM release 2. These experiments were conducted prior to the full preparation of the new release, so we only appended text from the original alignments of release 2 to this corpus. In total, the textual corpus used to train language models contains approximately 255 million words. These source data are described in~\cite{rousseau2014enhancing}.

\subsection{Experimental Results}
\label{subsec:T2expres}
In this section, we present the recent development on Automatic Speech Recognition~(ASR) systems that can be compared with the two previous releases of the TED-LIUM Corpus from 2012 and 2014. While the first version of the corpus achieved a WER of 17.4\% at that time, the second version decreased it to 11.1\% using additional data and Deep Neural Network~(DNN) techniques.
\subsubsection{TDNN} 

Our basis chain-TDNN setup is based on 6 layers with batch normalization, and a total context of (-15,12).
Prior tuning experiments on TED-LIUM release 2 showed us that the model did not improve beyond the dimension of 450. More than doubling the training data allows the training of bigger, and better, models of the same architecture as shown in Table \ref{T3tdnn}.\\

\begin{table}[h]
\centering
\caption{Tuning the hidden dimension of chain-TDNN setup on TED-LIUM release 3 corpus.}\label{T3tdnn}
\begin{tabular}{|@{\hspace{1em}} m{2cm} @{\hspace{1em}} | P{1.5cm} | P{1.5cm} | P{1.5cm} | P{1.5cm} | P{1.5cm} | P{1.5cm} |}
\hline
\multirow{2}{4em}{Dimension} & \multicolumn{2}{c|}{WER} & \multicolumn{2}{c|}{WER - Ngram} & \multicolumn{2}{c|}{WER - RNN} \\
\cline{2-7}
 & Dev & Test & Dev & Test  & Dev & Test \\
\hline
450 & 9.0\% & 9.1\% & 8.0\% & 8.4\% & 6.9\% & 7.3\% \\
\hline
600 & 8.7\% & 8.9\% & 8.0\% & 8.4\% & 6.6\% & 7.3\% \\
\hline
768 & 8.3\% & 8.6\% & 7.6\% & 8.1\% & 6.5\% & 7.0\% \\
\hline
1024 & 8.3\% & 8.5\% & 7.5\% & 8.0\% & 6.4\% & 6.9\%  \\
\hline
\end{tabular}
\end{table}

As part of experiments in tuning Kaldi models, it appeared that a form of L2 regularization could help to allow training for longer with less risk to overfit. This was implemented in Kaldi as the \texttt{proportional-shrink} option. Some tuning on TED-LIUM 2 data gave the best result for a value of 20.
All experiments presented in Table \ref{T3tdnn} were realized with this value to keep a consistent baseline. Aiming to reduce the WER even more, and with time constraints, we chose to train again the model with dimension 1024, with a proportional-shrink value of 10 (as we approximately doubled the size of the corpus). After RNNLM lattice-rescoring, the WER decreased to 6.2\% on the dev set and 6.7\% on the test.

\subsubsection{TDNN-F} As a final set of experiments, we tried the recently-introduced factorized TDNN approach, which again resulted in significant improvements in WER for both TED-LIUM release 2 and 3 corpora (Table \ref{T23tdnnf}).

\begin{table}[h]
\centering
\caption{Factorized TDNN experiments on TED-LIUM release 2 and 3 corpora.}\label{T23tdnnf}
\begin{tabular}{|@{\hspace{1em}} m{1cm} @{\hspace{1em}} |@{\hspace{1em}} m{3cm} @{\hspace{1em}} | P{1cm} | P{1cm} | P{1cm} | P{1cm} | P{1cm} | P{1cm} |}
\hline
\multirow{2}{4em}{Corpus} &\multirow{2}{4em}{Model} & \multicolumn{2}{c|}{WER} & \multicolumn{2}{c|}{WER - Ngram} & \multicolumn{2}{c|}{WER - RNN} \\
\cline{3-8}
& & Dev & Test & Dev & Test  & Dev & Test \\
\hline
r2 & TDNN-F - 11 layers - 1280/256 - ps20 & 8.5\% & 8.3\% & 7.8\% & 7.8\% & 6.8\% & 6.8\% \\
\hline
r3 & TDNN-F - 11 layers - 1280/256 - ps10 & 7.9\% & 8.1<\% & 7.4\% & 7.7\% & 6.2\% & 6.7\% \\
\hline
\end{tabular}
\end{table}

\section{Experiments with Fully Neural End-to-end ASR System}
\label{sec:e2eexp}
We also conducted experiments to evaluate the impact of adding data to the training corpus in order to build a neural \ee ASR. The system with which we experimented does not use a vocabulary to produce words, since it emits sequences of characters.
\subsection{Model Architecture}
\begin{figure}[!htbp]
\begin{center}
\includegraphics [scale=0.5]{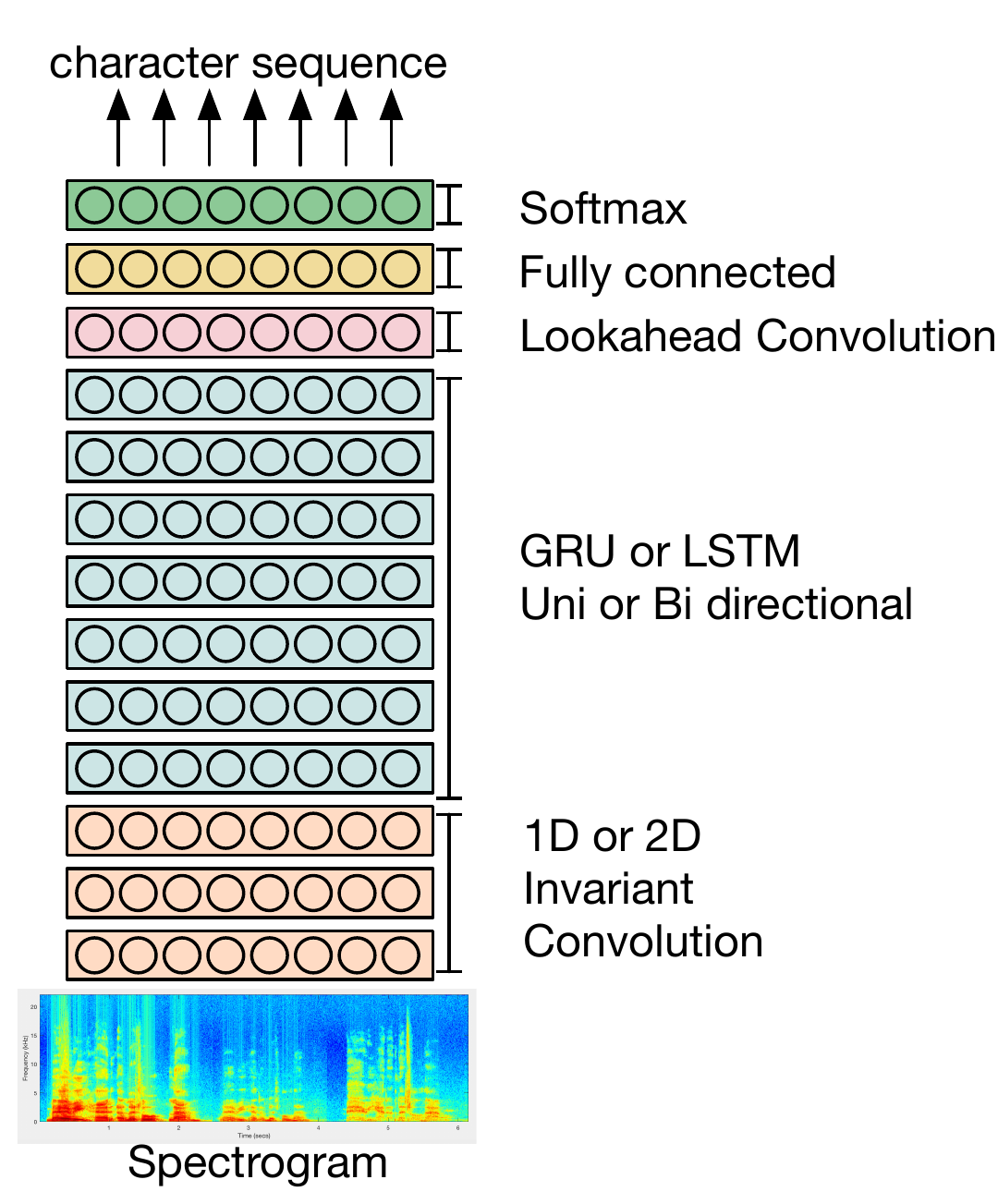}
\caption{\label{DeepSpeech} Deep Speech 2 -like end-to-end architecture for speech recognition.}
\end{center}
\end{figure}

The fully \ee architecture used in this study is similar to the Deep~Speech~2 neural ASR system proposed by Baidu in
~\cite{add23}.
This architecture is composed of $nc$ convolution layers (CNN), followed by $nr$ uni or bidirectional recurrent layers, a lookahead convolution layer~\cite{wang2016lookahead}, and one fully connected layer just before the softmax layer, as shown in Figure~\ref{DeepSpeech}.
The system is trained \ee by using the CTC loss function~\cite{graves2006connectionist}, in order to predict a sequence of characters from the input audio. 
In our experiments, we used two CNN layers and six bidirectional recurrent layers with batch normalization as mentioned in~\cite{add23}. 
Given an utterance $x^{i}$ and label $y^{i}$  sampled from a training set \
$X ={(x^{1}, y^{1}), (x^{2}, y^{2}), . . .}$, 
the RNN architecture has to train to convert an input sequence  $x^{i}$ into a final transcription $y^{i}$s. For notational convenience, we drop the superscripts and use $x$ to denote a chosen utterance and $y$ the corresponding label.
The RNN takes as input an utterance $x$ represented by a sequence of log-spectrograms of power normalized audio clips, calculated on 20ms windows. As output, all the characters $l$ of a language alphabet may be emitted, in addition to the space character used to segment character sequences into word sequences (space denotes word boundaries) and a \textsl{blank} character useful to absorb the difference in a time series length between input and output in the CTC framework.
The RNN makes a prediction $p(l_{t}|x)$ at each output time step $t$.  
At test time, the CTC model can be coupled with a language model trained on a large textual corpus. A specialized beam search CTC decoder~\cite{hannun2014first} is used to find the transcription $y$ that maximizes:
\begin{equation}
Q(y)=log(p(l_{t}|x)) + \alpha log(pLM(y)) + \beta wc(y)
\end{equation}
where wc(y) is the number of words in the transcription $y$. The weight $\alpha$ controls the relative contributions of the language model and the CTC network. The weight $\beta$ controls the number of words in the transcription.

\subsection{Experimental Results}
Experiments were made on the \textsl{legacy} distribution of the TED-LIUM~3 corpus in order to evaluate the impact on  WER of training data size for an end-to-end speech recognition system inspired by Deep Speech~2. In these experiments, we used an open source Pytorch implementation\footnote{\url{https://github.com/SeanNaren/deepspeech.pytorch}}.

\begin{figure}[h]
    \centering
\begin{tikzpicture}
\begin{axis}[
	height=6cm,
	width=6cm,
	grid=major,
	xlabel={Training data size (hours)},
	ylabel={Word Error Rate},
	name=wer,
	anchor=east
]
\addplot coordinates { 
	(205,28.1)    (290,23.1)   (452,20.3)
};

\addplot coordinates { 
	(205, 25.9) (290, 19.7) (452, 17.4) 
	};

\addplot coordinates {  
	(205,18.5)  
	(290,15.5)   (452,13.7)
};

  \end{axis}

\begin{axis}[legend pos=outer north east,
	height=6cm,
	width=6cm,
	grid=major,
	xlabel={Training data size (hours)},
	ylabel={Character Error Rate},
	name=cer,
	at={($(wer.east)+(1.5cm,0)$)},
	anchor=west,
	legend style={
			at={(-0.55, -0.38)},
			anchor=west}
]
\addplot coordinates{
	(205, 9.29)    (290, 8.32)   (452, 7.43)
};

\addplot coordinates { 
	(205, 8.69) (290, 7.43) (452, 6.56)
	};

\addplot coordinates{
	(205,7.52)    
	(290,6.93)   (452,6.09)
};
\legend{Greedy, Greedy+augm.,Beam+augm.}

\end{axis}

\end{tikzpicture}
\label{fig:e2eResults}
\caption{Word error rate (left) and character error rate (right) on the TED-LIUM~3 \textsl{legacy} test data for three \ee configurations according to the training data size.}
\end{figure}
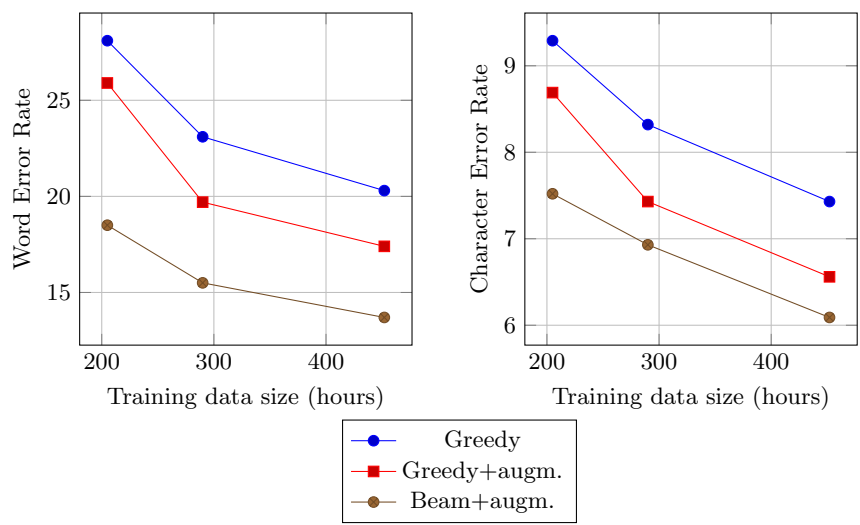

Three training datasets were used: TED-LIUM~2 with original alignment (207h of speech), TED-LIUM~2 with new alignment (290h), and TED-LIUM~3 (452h), as presented in section~\ref{subsec:align} and section~\ref{subsec:T3legacy}. They correspond to the three possible abscissa values (207, 290, 452) in figure~\ref{fig:e2eResults}.
For each training dataset, the ASR tuning and the evaluation were respectively made on the TED-LIUM release~2 development and test dataset, similar to the experiments presented in section~\ref{subsec:T2expres}.
Figure~\ref{fig:e2eResults} presents results in both WER (left side), and Character Error Rate (CER, right side) on the test dataset. Evaluation in CER is interesting because the \ee{} ASR system is trained to produce sequences of characters, instead of sequences of words.

For each training dataset, three configurations have been tested: 
\begin{itemize}
\item the \textsl{Greedy} configuration, in blue in Figure~\ref{fig:e2eResults} that consists of evaluating sequences of characters directly emitted from the neural network by gluing all the characters (including spaces to delimit words);
\item the \textsl{Greedy+augmentation} configuration, in red, which is similar to the Greedy one, but in which each training audio samples is randomly perturbed in gain and tempo for each iteration~\cite{ko2015audio};
\item the \textsl{Beam+augmentation} configuration, in brown, achieved by applying a language model through a beam search decoding on the top of the neural network hypotheses using the Greedy+augmentation configuration. This language model is the \textsl{cantab-TEDLIUM-pruned.lm3} provided with the Kaldi TEDLIUM recipe.
\end{itemize}

As expected, the best results in WER and CER are achieved by the \\
\textsl{Beam+augmentation} configuration, with a WER of 13.7\% and a CER of 6.1\%. Regardless of the configuration, increasing training data size significantly improves the transcription quality: for instance, while the Greedy mode reached a WER of 28.1\% with the original TED-LIUM~2 data, it reaches 20.3\% with TED-LIUM~3 .
We can observe that with TED-LIUM~3, the \textsl{Greedy+augmentation} configuration gets a lower WER than the \textsl{Beam+augmentation} one when trained with the original TED-LIUM~2 data. This shows that increasing the training data size for the pure \ee{} architecture  offers a higher potential for WER reduction than using an external language model in a beam search decoding.

\section{Experiments with the \textsl{Speaker Adaptation} Distribution}
\label{sec:SAexp}

In this section, we present results of speaker adaptation experiments on the adaptation version of the corpus described in Section~\ref{sec:adapt-corp}.
In this series of experiments, we trained three pairs of AMs. In each pair, we trained a speaker-independent (SI) AM and a corresponding speaker adaptive trained (SAT) AM. We explore two standard adaptation techniques: (1) i-vectors for a TDNN-LSTM and (2) feature space maximum linear regression (fMLLR) for a GMM-HMM and a TDNN-LSTM.
%
The Kaldi toolkit \cite{Povey_ASRU2011} was used for these  experiments.
First, we trained two GMM-HMM AMs on 39-dimensional features MFCC-39 (13-dimensional Mel-frequency cepstral coefficients
(MFCCs) with $\Delta$ and $\Delta\Delta$): (1) a SI AM and (2) a SAT model with fMLLR.
Then, we trained four TDNN-LSTM AMs.
All TDNN-LSTM AMs have the same topology,  described in~\cite{peddinti2018low},  and differ only in the input features. They were trained  using LF-MMI criterion~\cite{Povey2016PurelySN} and  3-fold reduced frame rate.
For the first SI TDNN-LSTM AM, 40-dimensional MFCCs without cepstral truncation (hires MFCC-40) were used as the input into
the neural network.
For the corresponding SAT model, i-vectors were used (as in the standard Kaldi recipe).
For the second  SI TDNN-LSTM AM, MFCC-39 features (the same as for the GMM-HMM) were used, and the corresponding SAT model was trained using fMLLR adaptation.
%
The 4-gram pruned LM was used for the evaluation\footnote{This  LM  is similar to the "small" LM trained with the pocolm toolkit, which is  used in the Kaldi \textit{tedlium s5\_r2}  recipe.  
The only difference is that we 
modified a training  set by adding  text data from TED-LIUM 3 and 
removing from it those data, that present in our test and development sets (from the adaptation corpus).}. 
Results in terms of WER are presented in Table\ref{tab:adapt-res}.


\begin{table}[h]
\centering
\caption{\label{tab:adapt-res}{Speaker adaptation results for the speaker adaptation task (on the corpus described in Section~\ref{sec:adapt-corp}. \textit{MFCC-39} denotes 13-dimensional MFCCs appended with $\Delta$  and $\Delta\Delta$; \textit{hires MFCC-40} denotes 40-dimensional MFCCs without cepstral truncation).}}
\centerline{\renewcommand{\tabcolsep}{2.0mm}
\begin{tabular}{|l|l|c|c|}
\hline
Model & Features & WER,\% -- Dev. &  WER,\% -- Test \\ \hline
GMM SI   & MFCC-39 & 20.69  & 18.02 \\ \hline 
GMM SAT  & MFCC-39 -- fMLLR & 16.47 & 15.08 \\ \hline 
TDNN-LSTM SI & hires MFCC-40 &7.69 & 7.25 \\  \hline 
TDNN-LSTM SAT & hires MFCC-40 $\oplus$ i-vect & 7.12 & 7.10\\  \hline 
TDNN-LSTM SI &  MFCC-39 & 8.19 & 7.54 \\  \hline 
TDNN-LSTM SAT &  MFCC-39 -- fMLLR  & 7.68 & 7.34\\  \hline  
\end{tabular}
}
\end{table}
\section{Discussion and Conclusion}
In this paper, we proposed a new release of the TED-LIUM corpus, which doubles the quantity of audio with aligned text for acoustic model training. We showed that increasing this training data reduces the word error rate obtained by a state-of-the-art HMM-based ASR system very slightly, passing from 6.8\% (release 2) to 6.7\% (release 3) on the \textsl{legacy} test data (and from 6.8\% to 6.2\% on the \textsl{legacy} dev data). To measure the recent advances realized in ASR technology, this word error rate can be compared to the 11.1\% reached by such a state-of-the-art system in 2014~\cite{rousseau2012ted}.
We were also interested in emergent neural end-to-end ASR technology, known to be very voracious in training data. We noticed that without external knowledge, \textsl{i.e.} by using only aligned audio from TED-LIUM~3, such technology reaches a WER of 17.4\%, which is exactly the WER reached by state-of-the-art ASR technology in 2012 with the TED-LIUM~1 training data. Assisted by a classical 3-gram language model used in a beam search on top of the end-to-end architecture, this WER decreases to 13.7\% with the TED-LIUM~3 training data, while with the TED-LIUM~2 training data the same system reached a WER of 20.3\%. Increasing training data composed of audio with aligned text  for this kind of ASR architecture still seems very important in comparison to the HMM-based ASR architecture that reaches a plateau on such TED data, with a low WER of 6.7\%.
Finally, we propose a new data distribution dedicated to experimenting on speaker adaptation, and propose some results that can be considered as a baseline for future work.

\subsubsection*{Acknowledgments.}
This work was partially funded by the French ANR Agency through the CHIST-ERA M2CR project, under the contract number ANR-15-CHR2-0006-01, and by the Google Digital News Innovation Fund through the \textsl{news.bridge} project.
\bibliographystyle{splncs04}
\bibliography{biblio}
\end{document}